\documentclass{article}

    \PassOptionsToPackage{numbers}{natbib}

    \usepackage[preprint]{neurips_2023}

\usepackage[utf8]{inputenc} 
\usepackage[T1]{fontenc}    
\usepackage{hyperref}       
\usepackage{url}            
\usepackage{booktabs}       
\usepackage{amsfonts}       
\usepackage{nicefrac}       
\usepackage{microtype}      
\usepackage{xcolor}         
\usepackage{graphicx}
\usepackage{algorithm}
\usepackage{algpseudocode}
\graphicspath{ {./plots/} }
\title{A Versatile Hub Model For Efficient Information Propagation And Feature Selection}

\author{%
  Zhaoze Wang$^{1,2}$\thanks{Zhaoze initiated this research during his visiting period at Shenzhen Technology University and finalized while undertaking Master's studies at University of Pennsylvania} \quad Junsong Wang$^{1}$\\
  $^1$Shenzhen Technology University \quad $^2$University of Pennsylvania\\
  \texttt{zhaoze@seas.upenn.edu} \quad \texttt{wangjunsong@sztu.edu.cn} \\
}

\begin{document}

\maketitle

\begin{abstract}
     Hub structure, characterized by a few highly interconnected nodes surrounded by a larger number of nodes with fewer connections, is a prominent topological feature of biological brains, contributing to efficient information transfer and cognitive processing across various species. In this paper, a mathematical model of hub structure is presented. The proposed method is versatile and can be broadly applied to both computational neuroscience and Recurrent Neural Networks (RNNs) research. We employ the Echo State Network (ESN) as a means to investigate the mechanistic underpinnings of hub structures. Our findings demonstrate a substantial enhancement in performance upon incorporating the hub structure. Through comprehensive mechanistic analyses, we show that the hub structure improves model performance by facilitating efficient information processing and better feature extractions.
\end{abstract}

\section{Introduction}

Topology plays a crucial role in determining the dynamics of both biological neural networks (BNNs) and artificial neural networks (ANNs). For ANNs, the topology and weight distribution during initialization are critical factors in determining the speed of convergence and the final states of the network \citep{pmlr-v9-glorot10a, pmlr-v28-sutskever13}. In the case of BNNs, it has been well-established that both functional and anatomical brain networks exhibit modularity \citep{10.3389/fnins.2010.00200,doi:10.1146/annurev-psych-122414-033634}, small-worldness \citep{Sporns2004, doi:10.1177/1073858406293182}, scale-free or log-normal degree distribution \citep{doi:10.1073/pnas.1111738109, 10.1093/cercor/bhl149, VANDENHEUVEL2008528, Bullmore2012}, and hub structures \citep{vandenHeuvel15775, VANDENHEUVEL2013683, Power2013, Bullmore2009, Bullmore2012}. Research in network neuroscience has demonstrated a strong correlation between cognitive functioning \citep{10.1371/journal.pcbi.1000395, Hearne2016}, information integration and propagation \citep{10.1371/journal.pone.0046497, vandenHeuvel14489}, and cognitive disorders \citep{10.1093/brain/awu132} with the topology of biological brains. These findings underscore the significance of topology in understanding and modeling neural networks.

Recent advancements in computational neuroscience have frequently employed RNNs as a means to study biological brains \citep{10.1371/journal.pcbi.1004792, Yang_2020, Pandarinath2018, BARAK20171}. Previous research has harnessed the capabilities of RNNs to investigate the emergence of spatial navigation \citep{cueva2018emergence}, decision-making, neuro-coding \citep{BARAK20171}, and sensorimotor learning \citep{Sussillo2009-sn}. Studies have also been proposed to load functional connectivity matrices to RNN to study the similarities as well as differences between ANNs and BNNs \citep{10.1371/journal.pcbi.1010639}. These in-silico simulations allow for a precise examination of the emergence of neuronal activation patterns and wiring patterns, which are difficult to obtain through in-vivo experiments. Therefore, training RNNs to perform cognitive tasks may provide insights into neuroscience. The adoption of biologically observed patterns may also improve the development of ANNs, as these features could be the result of natural selection and represent genetically optimized solutions.

When simulating BNN using ANN, a three-layer RNN is typically used \citep{10.1371/journal.pcbi.1004792, Rajan2016}. The first layer is a linear layer that serves to simulate the input signal to the simulated brain region. This input layer could either project signal to the entire hidden layer or a subset of hidden layer neurons depending on the experimental settings and objectives. The second layer is a randomly connected hidden layer, which serves as a simplified representation of the brain. It is often generated as a randomly connected sparse network, with weights following a Gaussian or uniform distribution. The last layer, denoted as the readout layer, maps the recurrent states into the targeting labels. Connection weights in the three layers could be updated using gradient descent \citep{10.1371/journal.pcbi.1004792}, intrinsic plasticity rules \citep{SCHRAUWEN20081159}, synatic plasticity rules \citep{SOLTOGGIO201848}, or remain unchanged \citep{10.1371/journal.pcbi.1010639, Jaeger2001TheechoST}.

Traditional RNN configurations, despite their simplicity and ease of implementation, may fall short in capturing the full range of biological characteristics of brain systems. It has been recognized that RNN architecture, particularly the hidden layer, significantly impacts convergence and should be carefully considered when modeling the brain. To achieve greater biological realism, several adaptations have been proposed. Specifically, \citet{10.1371/journal.pcbi.1004792} and \citet{10.1371/journal.pcbi.1003263} suggested to balance excitatory and inhibitory synapses in the hidden layer, while \citet{10.1162/netn_a_00082} have adopted the modular topology to the hidden layer.

In this work, we introduced the hub structure to the hidden layer initialization. Hub structure, also known as rich-club organization, denotes the topological attributes that biological brains contain many low-degree peripheral neurons \citep{VANDENHEUVEL2013683, 10.1371/journal.pcbi.1005989} and a few highly connected hubs for efficient information processing and propagation \citep{vandenHeuvel15775}. Our proposed hub structure generation method is flexible and broadly applicable to various RNNs, with a mathematical formalization that can also fit experimental data. We then applied the proposed hub structure on the echo state network (ESN) \citep{Jaeger2001TheechoST}, a special type of RNN in the reservoir computing (RC) paradigm that allows synaptic weight to remain unchanged during the course of training to emphasize the topological differences. Upon integrating our proposed hub model on ESN, our model demonstrates substantial improvements in prediction accuracy on several time-series forecasting tasks and a classification task. We provided extensive mechanistic analysis and identified that hub neurons could efficiently regulate the network states. Alongside this, our hub structure demonstrated enhanced feature extraction capabilities. Finally, a preference for peripheral neurons over hub neurons during prediction is observed, denoting an emergence of functional hierarchy after adopting the hub structure.

\section{The Hub Model}

\subsection{Motivation} 

The brain topology is often characterized by its modular organization, small-world characteristics, log-normally distributed nodal degrees, and the existence of neuronal hubs. These features are interlinked, each underscoring different facets of the brain's organization. For the hub structure, it hints at modularity, with each hub potentially acting as the central node for its associated peripheral neurons. It also implies a long-tailed distribution of nodal degree, as it requires only a few highly connected neurons, resulting in a log-normal degree distribution. Furthermore, hub neurons themselves form a small network, which significantly reduces the average nodal length, indicative of small-world properties. Therefore, in our research, we focus on the hub structures as they often imply the presence of the other three properties. Additionally, most recent neuroscience and cognitive research suggest this rich-club organization correlates with individuals' ability on several cognitive tasks \citep{Bertolero2018}.

Several factors have been proposed to account for the emergence of hub structures in networks. Wiring cost considerations, which balance network efficiency and metabolic expenses and spatial volume, are known to shape hub structures \citep{Bullmore2012}. Neurogenetic elements, encompassing genetic predispositions, developmental processes, and neuronal migration and differentiation, are also crucial in hub neuron formation \citep{VANDENHEUVEL2013683}. Furthermore, Hebbian learning principles, encapsulated by the postulate "neurons that fire together wire together," may foster hub structure development by reinforcing commonly used connections and further facilitating the creation of highly connected hub neurons.

\subsection{Hub generation} 

In this study, we consider neurogenetic factors \citep{VANDENHEUVEL2013683} and wiring cost \citep{Bullmore2012} as the main contributors to the emergence of hub structure during network initialization. Our objective is to investigate the impact of hub topology on network dynamics. Consequently, we have excluded Hebbian factors from the analysis, as their inclusion would necessitate a more intricate discussion on synaptic adaptations. Our proposed method involves first constructing a densely connected weight matrix \( W \). Then a pruning probability matrix \( P_{prune} \) is constructed based on the constraining factors, with each of its entries \( p_{ij} \) representing the likelihood of the corresponding edge \( e_{ij} \) in \( W \) be removed until \( W \) meets the pre-defined sparsity level.

The wiring cost primarily imposes constraints on neuron density, synapse distance, and axonal projections cross section diameters \citep{Bullmore2012}. In our proposed model, the distance factor is considered and termed as distance constraint (DC). Constraints imposed by neurogenesis factors are denoted as neurogenetic constraints (NC). Lastly, we introduced a regularization term \( R \) to allow for a smooth transition from hub network to random network.

To construct DC, each neuron \( n_i\ |\ i \in N\), where \( N \) is the network size, is randomly assigned a 3D coordinate \( Q_{i} = Q(x, y, z)\ |\ x, y, z \sim \mathcal{N}(0, 1) \) at initialization (as depicted in Fig. 1a). Following which, the DC matrix \( C_{d} \) is constructed using the neuron coordinates, with each entry \( dc_{ij} \in C_{d} \) is the Euclidean distance between node \(i\) and \(j\), as shown in Eq.1.

\begin{figure}
  \centering
  \includegraphics[width=\textwidth]{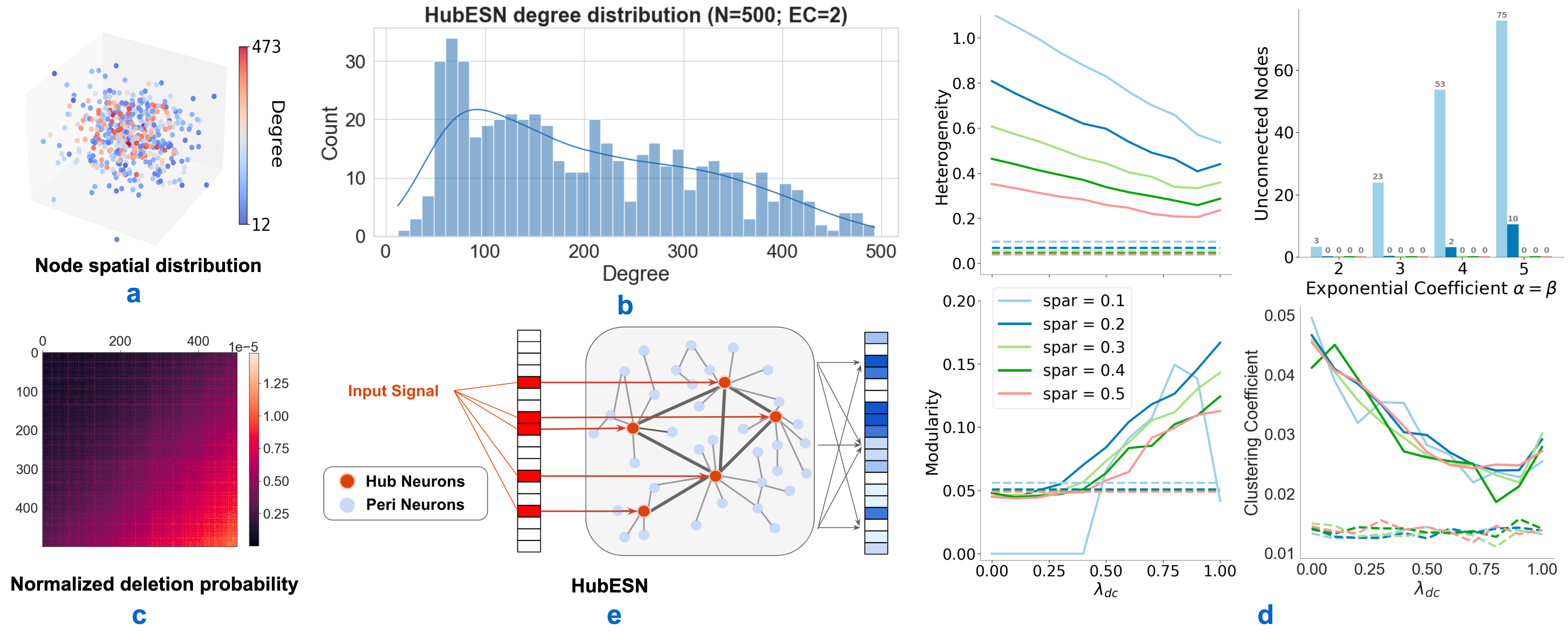}
  \caption{\textbf{a-d)} Topological properties of a 500-neuron hub network. Unless otherwise specified, \( \alpha = \beta = 2 \). \textbf{a)} A 3-d coordinate set was generated to construct DC, nodes in the cluster center tend to have higher degrees. \textbf{b)} The network (N=500; spar=0.2) degree follows a log-normal distribution. \textbf{c)} Normalized deletion probability \( P_{prune} \). DC contributes to the lattice-like pattern and NC contributes to the gradient-like pattern along the diagonal in deletion probability matrix \( P_{prune} \). \textbf{d)} The top-left, bottom-left, and bottom-right panels show heterogeneity, modularity, and clustering coefficient in relation to the \( \lambda_{dc} \), respectively. The results displayed are the means of five repetitions of the experiment. \( \lambda_{nc} \) were set to \( 1 - \lambda_{dc} \). Colors indicate sparsity, dashed lines represent random networks, and solid lines denote hub networks. The top-right panel depicts numbers of unconnected nodes under different exponential coefficient values, which is undesirable and should be avoided. \textbf{e)} HubESN architecture, signals are injected into the hub neurons. }
\end{figure}

\begin{equation}
dc_{ij} = d(Q_i, Q_j)\ |\ \forall dc_{ij} \in C_d;\ i,j \in N
\end{equation}
\begin{equation}
nc_{ij} = i + j\ |\ \forall nc_{ij} \in C_n;\ i,j \in N
\end{equation}
\begin{equation}
p_{ij} = \frac{\lambda_{dc} dc_{ij}^{\alpha} + \lambda_{nc} nc_{ij}^{\beta} + \lambda_{reg} R}{\sum_{k, l \in N} (\lambda_{dc} dc_{kl}^{\alpha} + \lambda_{nc} nc_{kl}^{\beta} + \lambda_{reg} r_{kl})}\ |\ \forall p_{ij} \in P_{prune};\ i,j \in N
\end{equation}

We consider the NC to encompass a confluence of genetic influences, developmental trajectories, neuronal migration patterns, and neuron generation timing. In an effort to construct a simplified yet representative assumption, we define each entry \( nc_{ij} \) in \( C_n \) to be the sum of the indices \( i \) and \( j \), as indicated in Eq.2. We acknowledge that this assumption simplifies the complex nature of the underlying processes. However, this construction method for NC remains both efficient and effective in generating the hub structure. Additionally, it fosters a symmetrical connection pattern for both pre-synaptic and post-synaptic weights.

Finally, \( \alpha \) and \( \beta \) are exponential coefficients (EC) that serve to raise the power of \( C_d \) and \( C_n \) to fit a log-normal nodal distribution (Fig. 1b). \( \lambda_{dc} \) and \( \lambda_{nc} \) are scaling coefficients (SC) for \( C_d \) and \( C_n \), while \( \lambda_{reg} \) is the SC for the regularization term \( R \). Entries \( r_{ij} \in R\) are random numbers following the same distribution as entries in \( W \), such that increasing \( \lambda_{reg} \) will gradually transform hub network to random network. As demonstrated in Eq.3, both SC and EC are applied to the constraints, and then the results are combined to generate the pruning probability matrix \( P_{prune} \). The entries in \( P_{prune} \) are normalized such that they sum up to 1 (Fig. 1c). \( P_{prune} \) is subsequently used to prune the densely connected weight matrix \( W \) to the pre-defined sparsity.

Fig. 1a shows the impact of DC that the centered neurons are more likely to be connected and therefore have higher nodal degrees, which is computed as the sum of in-degree and out-degree. Fig. 1c illustrates the normalized  \( P_{prune} \), where DC contributes to lattice-like patterns, and the NC contributes to the gradient pattern along the diagonal. Brighter colors indicate a higher probability of deletion. While NC fosters a clear, scale-free distributed pruning probability along the diagonal, a potential drawback is that it might also introduce unconnected neurons by overly emphasizing the deletion of edges that connect peripheral neurons. DC alleviates this unconnected neuron problem by encouraging connections between peripheral neurons and their nearest neighbors (Suppl. 1).

It is pertinent to mention that this model is not exclusively designed for RNN initialization; it may also function as a theoretical framework to accommodate experimental data. It could potentially be used to fit on experimentally collected functional connectivity matrices by adjusting the SCs and the ECs. This adaptability makes the model a potential framework for generalizing and simplifying the complexity of experimentally collected data.

\subsection{Topological features and parameter choices}
The incorporation of a hub structure precipitates numerous topological changes within the network. We use network heterogeneity, modularity, and clustering coefficient under difference network sparsity to assess the topological difference after adopting the hub structure (Fig. 1d). We use the coefficient of variation (CV) of the nodal degree to quantify network heterogeneity \citep{Litwin-Kumar2017-uj, JU201339} (Suppl. 3.1). The network modularity is measured by first partitioning the network via the Louvain method \citep{Blondel_2008}, followed by computation using Newman’s community detection algorithm \citep{doi:10.1073/pnas.0601602103} (Suppl. 3.2). The clustering coefficient \citep{Saram_ki_2007} is approximated by converting the network into weighted bidirectional graphs (Suppl. 3.3). Our findings reveal that the hub network manifests a significantly elevated CV compared to its random counterpart, with NC contributing more to heterogeneity than DC (Fig. 1d, upper left). Furthermore, we noted a surge in network modularity along with a subtle augmentation in the clustering coefficient. This suggests that the introduction of a hub structure encourages the development of sub-networks and network heterogeneity.

On top of providing a versatile mathematical model of hub structure, we aim this study to provide a mechanistic explanation of hub structure. Therefore, we have opted for the simplest parameters wherever possible throughout this paper, in order to streamline the discussion and highlight the distinctions between a standard random network and a hub network. We use the most basic ECs, \( \alpha = \beta = 2 \), as they provide a suitable balance between model complexity and interpretability. We have confirmed that the selection of EC does not significantly influence the network's topological properties, nor does it affect our final conclusions (refer to Supplementary Material 2).

While the choice of EC does not substantially impact network topological properties, we observed notable differences between various SC values. It is important to note that \( \lambda_{dc} \), \( \lambda_{nc} \), and \( \lambda_{reg} \) are scale-invariant, as they will be subsequently normalized. Their values only represent their relative importance compared to one another. As such, we let \( \lambda_{nc} + \lambda_{dc} + \lambda_{reg} = 1 \). As we aim to contrast the difference between random network and hub network, \( \lambda_{reg} \) is set to 0 throughout this paper so that the network will demonstrate strongest hub features. However, the distance constraint (DC) decreases heterogeneity and clustering coefficient while increasing modularity (Fig. 1d). To streamline the discussion, we chose \( \lambda_{dc} = \lambda_{nc} = 0.5 \) for the rest of the paper.

\section{Hub echo state network}
\subsection{Echo State Network}
The Echo State Network \citep{Jaeger2001TheechoST, NIPS2002_426f990b} is adopted to investigate the effects of hub structures on network performance, dynamics, and mechanisms. The architecture has been frequently used in the investigation of machine learning and neuroscience \citep{10.1371/journal.pcbi.1010639, 10.1162/netn_a_00082, Sussillo2009-do} The input and hidden layers of ESNs remain unchanged during training, serving only to map input signals to a higher dimension before being translated to output by the linear regression readout layer. By maintaining a consistent synaptic configuration, we ensure the preservation of the hidden layer topological features. This clarity enables a more accurate attribution of performance improvements and mechanisms to the network topology rather than synaptic updates.
\begin{equation}
s(t+1) = f(W^{in} u(t+1) + W^{rec} s(t))
\end{equation}

The update function of ESN is shown in Eq.4, where \( s(t) \) is the hidden state at time \(t\), \( u(t) \) is the input for time \(t\);  \(f( \cdot) \) is the non-linear activation function, and \( W^{in} \) and \( W^{rec} \) are weight matrices for input and hidden layer, respectively. During training, the input \( u(t) \) is sequentially presented to the ESN. The ESN hidden states are stacked into a state matrix \( S=[s(0);s(1) ...; s(T)] \). At the end of the training epoch, the readout layer computes the output weights using the closed-form solution of linear regression, \( (S^T S)^{-1}S^TY \), where \( Y \) is the label vector. During testing, ESN behaves like a standard RNN that maps input to hidden states and then makes predictions at each timestep.

\subsection{HubESN}
Hub neurons, serving as central nodes within neuronal modules, are recognized in neuroscience research as crucial elements in enabling the efficient dissemination of information from various brain regions. In alignment with previous neuroscience findings that hub neurons facilitate communications across different brain regions, we assume that the input layer simulates the signals projected from other brain regions. The connection between input layer and hidden layer cannot be fully connected, as this would negate the concept of "information transfer between two brain regions". We assume only 10\% of the neuron receives input from other brain regions, denoted as \( r_{sig}=0.1 \).

We denote the ESN after incorporating hub structure as HubESN. For standard ESN, the network nodal degrees are homogeneous. Therefore, signals are injected randomly to 10\% of the nodes. For HubESN, we propose two methods of injecting signals: similar to ESN, inject signal randomly to 10\% of the neurons; or inject signal to the top 10\% of neurons with the highest number of connections. We denote the first method as HubESN-rand, and the latter as HubESN. As hub neurons are characterized by having more connections compared to other neurons, we assume signals are injected into hub neurons in the second setting. The illustration of HubESN is demonstrated in Fig. 1e, where red arrows indicate where signals were injected, red nodes represent hub neurons, and blue nodes represent peripheral neurons. The complete implementation procedure and parameter choice of HubESN, HubESN-rand, and ESN are included in Suppl. 5.

For weights \( w_{ij} \in W^{rec} \) of ESN, HubESN, and HubESN-rand, \( w_{ij} \sim \mathcal{N}(0, 1/3) \). We opt for \( tanh( \cdot ) \) as our activation function such that the hidden states will oscillate between \( (-1, 1) \). An additional imperative parameter for ESN is the spectral radius, which scales the maximum eigenvalue of \( W^{rec} \). As prior research identifies a spectral radius slightly less than 1 is required to guard the echo state property \citep{NIPS2002_426f990b, Lukoševičius2012}, we accordingly set it to 0.9. The network sparsity is set at 0.2, selected based on the that this value yields the most considerable disparity between the hub network and a random network (Fig. 1d). We use the spectral radius of 0.9 and sparsity of 0.2 consistently throughout this paper. Our conclusions remain unaffected by these parameter choices (Suppl. 4).

\section{Experiments}
We benchmarked HubESN and standard ESN using three standard tasks: Mackey-Glass \citep{Jaeger2001a} prediction task, nonlinear autoregressive moving average model (NARMA10) \citep{Jaeger2001a, LUKOSEVICIUS2009127, goudarzi2014comparative} prediction task, and MNIST \citep{7982291} written digit classification task. All three tasks are standard metrics to evaluate ESN performance. 

\begin{equation}
\frac{dx_{m}}{dt} = \frac{\beta_{m} x(t-1-\tau)}{1+(x_{m}(t-1-\tau))^{k}}-\gamma_{m} x_{m}(t-1)
\end{equation}

\begin{equation}
x_{n}(t) = \alpha_{n} x_{n}(t-1) + \beta_{n} x_{n}(t-1) \sum_{i=1}^{l} x_{n}(t-i) + \gamma_{n} u_{n}(t-l) u_{n}(t-1) + \delta_{n}
\end{equation}

The Mackey-Glass equation is a delayed differential equation exhibiting complex nonlinear dynamics, and was initially introduced as a model for physiological control systems with time delays. We employ this task to evaluate the model's capacity for retaining long memory and leveraging non-linearity to achieve accurate predictions. The Mackey-Glass time series is updated as Eq. 5. In alignment with previous literature \citep{Jaeger2001a, DBLP:journals/corr/GoudarziBLTS14}, the parameters \( \tau \), \( \gamma_{m} \), \( \beta_{m} \), \( k \) are set to \( 17 \), \( 0.1 \), \( 0.2 \), and \( 10 \), respectively. \( dt \) is set to \( 1 \) to discretize the equation. At each time step \( t \), \( x_n(t) \) is presented to the network, which updates its hidden states to \( s(t+1) \) and predicts \( x_n(t+1) \) using the information it has encoded.

The NARMA10, a tenth-order nonlinear difference equation, is employed in our study to evaluate the model's proficiency in capturing complex non-linear dependencies. In conjunction with the Mackey-Glass series, it serves to determine the versatility of the proposed model in predicting different types of time series. NARMA10 task is defined as Eq. 6., with \( l=10 \), \( \alpha_{n}=0.3 \), \( \beta_{n}=0.05 \), \( \gamma_{n}=1.5 \), \( \delta_{n}=0.1 \) in consistence with previous research \cite{goudarzi2014comparative}. An independent random variable \( u(t) \in [0, 0.5] \) is presented at each time step, and predictions are made based on the previous \( l = 10 \) inputs.

Finally, the MNIST task is employed to if the model performance improvements can be generalized to non-time series tasks. The network predicts labels of written digits by scanning each 28x28 image column-wise. During training, each image is input into the network over 28 time steps, with the one-hot label used for training the model at all 28 steps. During testing, the final label is determined by majority voting based on the network's predictions.

Unlike prior studies that use a fixed training sample size for ESN benchmarking, this study assesses models under various training sample sizes to allow for a more comprehensive evaluation of convergence patterns. We train models on differently-sized datasets and subsequently test on a fixed-size test set. The size of train set and test set are denoted as n\_train and n\_test in this paper. For the Mackey-Glass and NARMA10 tasks, models are trained on n\_train steps of signal and evaluated on their ability to predict the subsequent 2000 steps, with performance measured by root mean square error (RMSE). For the MNIST task, we randomly select n\_train images for training and measure classification accuracy on a test set of 3000 images. 

Due to the sensitivity of ESN to random initialization, each trial setting is repeated 100 times for Mackey-Glass and NARMA10, and 10 times for MNIST, to ensure measurement precision. To mitigate potential bias, the same dataset is provided to ESN, HubESN, and HubESN-rand models in each trial (refer to Suppl. 6 for replication specifics).

\section{Results}
\subsection{Performance}

Fig. 2 presents the performance of ESN, HubESN, and HubESN-rand. We use network size of N=1000 on Mackey-Glass and NARMA10 prediction tasks and a smaller network N=500 on MNIST tasks to reduce training time. The network testing loss for Mackey-Glass and NARMA10 prediction tasks decreases exponentially as n\_train increases, therefore in the lower plots of Fig. 2a and 2b, HubESN and HubESN-rand performances are presented as ratios with respect to ESN performance to facilitate a more lucid visual differentiation of ESN and HubESN performance trajectories.

On Mackey-Glass and NARMA10, when the number of training samples is small, while HubESN-rand is slightly worse than HubESN, both significantly outperform ESN. On Mackey-Glass prediction, within the substantial training set size range of 600-2000, HubESN reduces RMSE by more than 37\%. At n\_train = 1200, HubESN reduces RMSE by 57\%. As the training set increases, ESN performance gradually approaches HubESN and HubESN-rand. Note that, however, ESN does not update weights in hidden and input layers; the improved performance is solely attributed to the readout layer. As the ESN readout layer predicts using linear regression, better performance on the same training set size itself suggests a more linearly separable hidden state.

Although ESN performance will gradually approach HubESN performance on time-series tasks, it does not apply to the classification tasks. On MNIST, we benchmarked the model on a wide range of train set sizes. Increasing the training set size did not improve ESN performance to match HubESN performance. Both ESN and HubESN performance stopped increasing after n\_train became greater than 12500, while the classification accuracy of HubESN remains greater than that of ESN for all training set sizes.

\begin{figure}
  \centering
  \includegraphics[width=\textwidth]{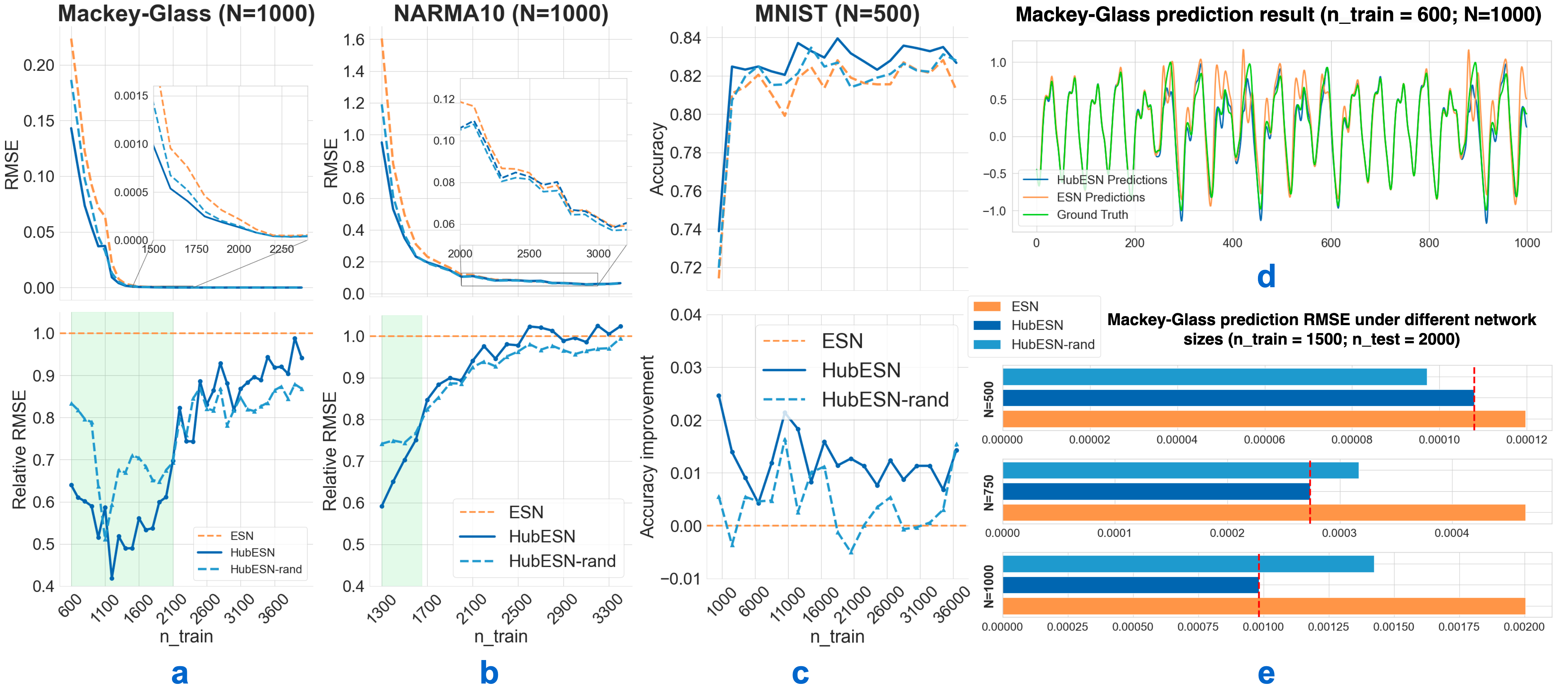}
  \caption{\textbf{a-c)} Testing performance of HubESN, HubESN-rand, and standard ESN with varying n\_train values on Mackey-Glass, NARMA10, and MNIST tasks. The line is solid if signals are input to hub neurons and dashed if to random neurons. Line hues differentiate network types. The top row exhibits absolute testing performances. For panel \textbf{a} and \textbf{b}, the second row shows RMSE relative to standard ESN for clarity of comparison. For panel \textbf{c}, the second row shows accuracy improvements. \textbf{d)} Sample prediction result (n\_test=1000) for Mackey-Glass. Predictions from HubESN (blue line) follows more closely to the ground truth than predictions from ESN (orange line) \textbf{e)} Given a fixed number of n\_train, the performance advantage of HubESN increases as the network size grows.}
\end{figure}

The size of an Echo State Network (ESN) also impacts its performance. While larger networks can theoretically accommodate more intricate dynamics, they also amplify the complexity of hidden states, complicating readout. In Fig. 2e, we depicted model performance differences on Mackey-Glass prediction when network sizes are N=500, N=750, and N=1000. As increasing network size introduces more complex reservoir dynamics, the readout layer requires larger training set to converge. Therefore, the superiority of HubESN becomes more pronounced with larger network sizes.

\begin{figure}
  \centering
  \includegraphics[width=\textwidth]{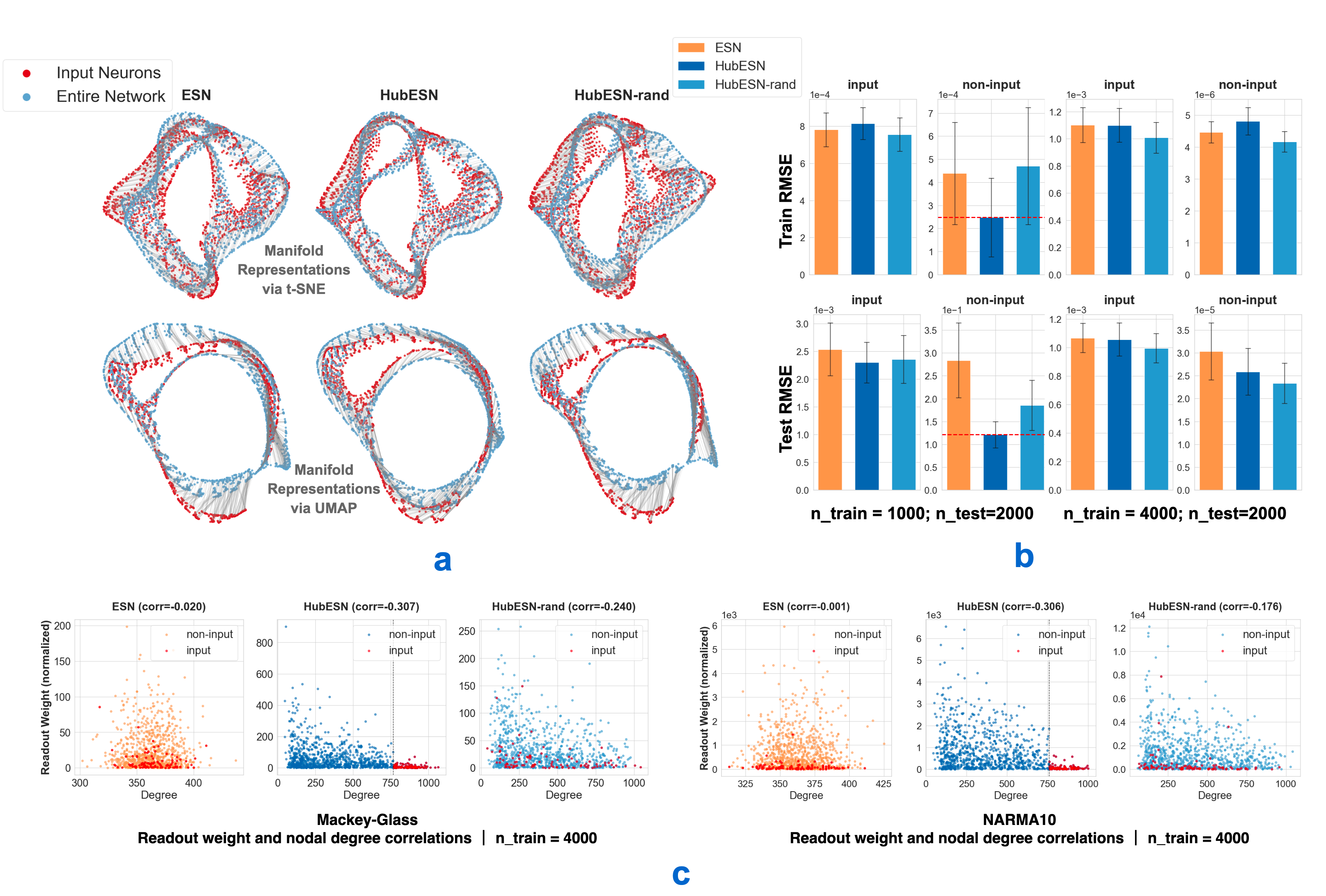}
  \caption{\textbf{a-c)} Visualizations derived from N=1000 size networks. \textbf{a)} Neural manifolds for the full network and input neurons only. Dimension reduced via t-SNE and UMAP on Mackey-Glass time series. Red and blue dots indicate input neuron states and entire network states respectively, with lines connecting temporally corresponding embedded states. \textbf{b)} Training and testing errors using only input neurons or non-input neurons to predict at n\_train of 1000 and 4000. Errorbars denote the SD of the measurements. \textbf{c)} An inverse correlation observed between readout weight and nodal degree is observed, regardless of the task (plots for MNIST are included in Suppl. 7).}
\end{figure}

\subsection{Hub neurons for efficient information propagation}
In the areas highlighted by green in Figures 2a and 2b, as well as throughout Figure 2c, the HubESN model outperforms both HubESN-rand and ESN. The sole distinction between HubESN and HubESN-rand lies in where the hidden layer receives input. We applied t-Distributed Stochastic Neighbor Embedding (t-SNE) \citep{JMLR:v9:vandermaaten08a} and Uniform Manifold Approximation and Projection (UMAP) \citep{mcinnes2020umap} to visualize the higher dimensional hidden layer states over time to investigate the mechanistic differences. The Mackey-Glass time series is utilized as it is structured in higher dimensions. The hidden layer states were recorded over time, and dimensionality reduction algorithms were applied to reduce hidden state dimensions and preserve their structure in lower dimensions. The neural manifold trajectories of input neurons only and the whole network are visualized in Fig. 3a. We use them The line segments connect the network states and their corresponding input neuron states.

In the ESN and HubESN-rand models, the overall network trajectories do not align closely with the trajectories of the input neurons, whilst the HubESN model shows a close correspondence between the network and input neuron state trajectories (Fig. 3a). This close alignment between the neural manifold trajectories of input neurons and the entire network signifies a more cohesive system within HubESN compared to HubESN-rand and ESN, hence implying an elevated significance of hub neurons in transmitting information throughout the network.

We further group the neurons into two subsets, those receiving direct input from the input layer (input neurons) and those that do not (non-input neurons), to investigate their roles. The models are run (without fitting the readout layer) on a smaller (n\_train=1000) and a larger (n\_train=4000) Mackey-Glass training set. The hidden layer states were grouped into input neuron states and non-input neuron states. We subsequently fit a linear regression model to both input neuron states and output neuron states and use them to predict the training and test sets. 

In low n\_train conditions (Fig. 3b, lower and upper left), input neurons across ESN, HubESN, and HubESN-rand exhibit similar prediction capabilities, while all models display higher testing loss than training loss (observe the y-axis scale difference). Conversely, HubESN outperforms in predicting through non-input neurons on both training and testing sets. As HubESN differs HubESN-rand only in its input injection location, and its network trajectory closely follows input neuron trajectories, we contend that HubESN improves performance through efficient information distribution. This efficiency elicits quicker response from a greater number of neurons, fostering rich feature generation and improved predictions. As the vast majority of neurons are non-input (90\% against 10\% input neurons), their effective use significantly augments overall performance.

When n\_train increases (Fig. 2b), non-input neurons show a marked improvement in prediction capabilities, implying their complex states require larger training samples to be accurately captured. Conversely, input neurons exhibit less prominent improvement. This trend underscores the importance of efficient information distribution from input neurons throughout the network.

This claim is further supported by MNIST performance, where HubESN consistently outperforms ESN and HubESN-rand regardless of training set size. MNIST task requires the model to produce as many correct labels as possible within a given time (n=28 for 28 columns). The delayed time dependency cannot be remedied by increasing the training set size. Therefore, this implies a greater advantage of hub structure on non-time series tasks where delayed feature representation cannot be learned.

\subsection{Emergence of rich features and functional hierarchy}
The efficient information distribution of hub neurons explains the superiority of HubESN over other models in MNIST tasks and at low n\_train levels and classification tasks. However, this does not fully account for the performance improvement of HubESN-rand across a broad range of n\_train values (Fig. 2a and 2b). We contend the hub structure itself also provides a better feature extraction ability. This can be supported by the fact that non-input neurons in HubESN-rand also have a lower prediction RMSE than ESN in both small and large n\_train values (Fig. 3b, lower panels). As HubESN-rand differs ESN only in its hidden layer topology, it shows an inherent effective feature extraction and generalization ability of hub structure.

Existing neuroscience research also suggested a functional difference between hub neurons and peripheral neurons. While hub neurons generally serve for information propagation and integration, peripheral neurons are typically specialized for certain types of processing. With the proposed hub structure, we examined if such functional specialization also emerges in HubESN and HubESN-rand during prediction.

The readout layer of ESN is a linear regression that maps the hidden states to output. Therefore the absolute readout weight can serve as an indicator of a neuron's importance to prediction. We have taken the fact that neurons can oscillate across different magnitudes into account. That is, neurons with smaller absolute oscillation magnitude may have larger weights in the readout layer. We normalized weights using average absolute magnitude across time to ensure an accurate reflection of a neuron's prediction importance. We stack the neuron degree and normalized weights into two vectors and compute their correlations. Fig. 3c reveals a negative correlation between degree and weight in both HubESN and HubESN-rand, indicating a preference for peripheral neurons in predictions. This pattern is not task-specific and is observed across all tasks (Suppl. 7). The preference for peripheral neurons in prediction indicates a role difference between hub neurons and peripheral neurons. Moreover, as this negative correlation exists in both HubESN and HubESN-rand, suggests that this functional difference between nodes with different degrees is self-emerged from hub structure. Finally, Fig. 3c also reflects the neurons that receive input (colored in red) have lower importance in prediction as they have smaller normalized weights, verifying the results in Fig. 3b that input neurons have higher RMSE than non-input neurons when training samples are sufficient.

\section{Discussion}
In this work, we present a biologically plausible hub model that is both versatile and amenable to control over various constraining factors of hub structure and nodal degree distributions. The flexibility of the proposed hub model allows for seamless transitions between different degrees of heterogeneity, modularity, and clustering coefficients. Our hub model can be readily applied to a range of RNNs to create biologically realistic RNNs, fit onto experimentally collected functional connectivity maps, and used to uncover underlying brain function mechanisms.

Moreover, as a machine learning model, our HubESN demonstrated significant performance improvements across numerous tasks (Fig. 2a-c), outperforming ESN when training data was limited and in classification tasks. Through extensive experimental analysis, we believe it is fair to conclude that the improved performance primarily attributes to the efficient information propagation ability of hub neurons and the better feature extraction ability of the hub structure. The preference for peripheral neurons during prediction aligns with neuroscience insights that peripheral neurons tend to have task-specific responses.

While we chose ESN architecture to ease the process of analyses, its unchanged synapse weights also limited us from deeper mechanistic investigations. For instance, in addition to information propagation, hub neurons may also function to integrate signals from peripheral neurons and play a critical role in multisensory integration. As ESN hidden layer synaptic weights are unchanged over time, it would be challenging for the simple linear regression layer to read out the integrated signal. Additionally, hub structure is also identified in a wide range of brain areas and may have different functions depending on their anatomical locations. Future research could apply our hub model to more advanced RNN structures, training on realistic coherence signals designed for specific cognitive tasks. Moreover, despite our conclusions being unaffected by selected EC and SC values, comprehensive future studies could cover a wider parameter space.

Overall, while the functioning mechanisms of both BNNs and ANNs function remain an active field of research, our hub model could serve as an efficient tool for bridging BNNs with ANNs. Our HubESN further validates the advantages of hub structure and hub neurons within ANNs.

\begin{ack}
We would like to acknowledge the assistance of OpenAI's language model, ChatGPT, which helped in paraphrasing and improving the clarity of this paper's presentation.
\end{ack}


\bibliography{references}

\newpage
\begin{suppl}
\setcounter{section}{0}
\setcounter{figure}{0}
\section{The impact of DC on unconnected neurons}
\begin{figure}[!h]
  \centering
  \includegraphics[width=\textwidth]{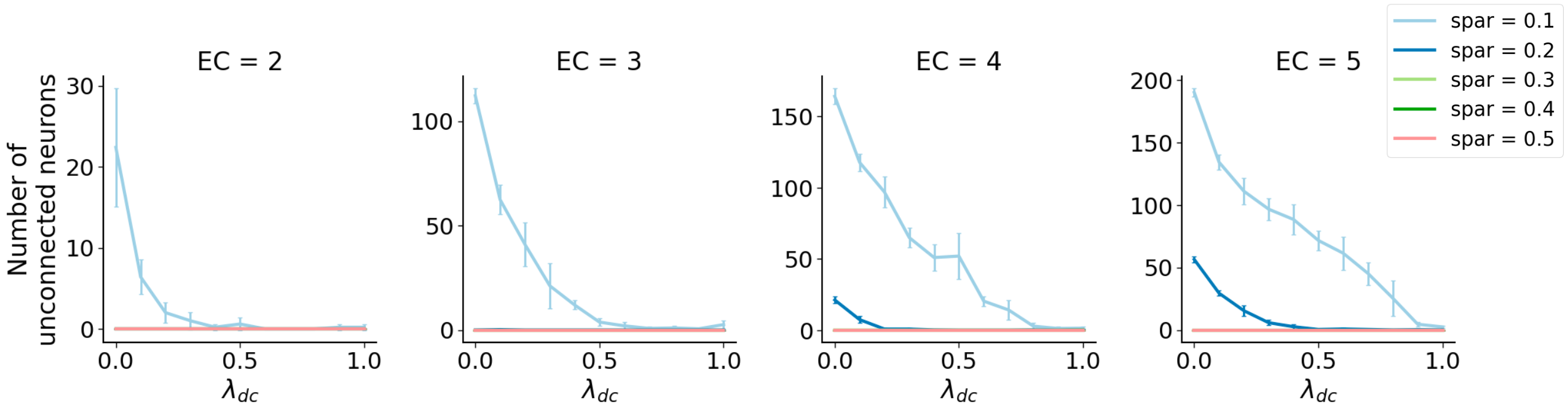}
  \caption{Number of unconnected neurons under different Exponential Coefficient (EC) values. Error bars denote the SD of the measurements. }
\end{figure}
We define neurogenetic constraints (NC) as Eq. 1. That is, the probability of an edge being deleted depends on the indices of the two neurons it connects. 

\begin{equation}
nc_{ij} = i + j\ |\ \forall nc_{ij} \in C_n;\ i,j \in N
\end{equation}

However, as it will be subsequently raised by the power of \(  \beta \) to fit a log-normal distribution, this setting may over-emphasizing deleting higher indices neurons. When \( \beta \) is high and network sparsity is low, NC may result in some neurons being unconnected to the network. This is biologically unrealistic and will also degrade network performance as the network cannot use all of its neurons to predict.

On the other hand, DC deletes edges based on the Euclidean distance between two neurons. It is prone to preserve the edges that connect neurons and their closest neighbors, thereby alleviating the unconnected neuron problem. As shown in Suppl. Fig. 1, increasing the value of \( \lambda_{dc} \), i.e., increasing the relative emphasis on DC constraint, will lower the number of unconnected neurons.

\section{The impact of EC}
\begin{figure}[!h]
  \centering
  \includegraphics[width=0.8\textwidth]{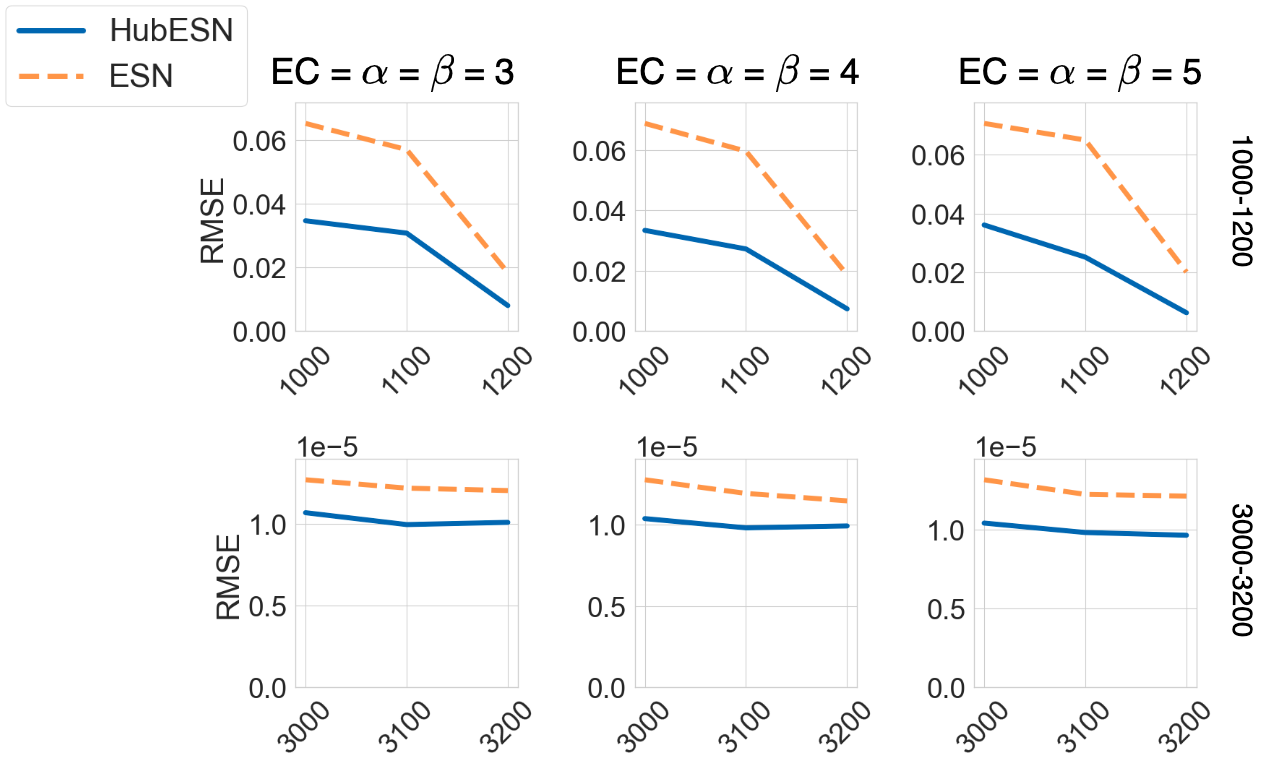}
  \caption{Testing RMSE for ESN and HubESN when EC = 3, 4, and 5.}
\end{figure}

As highlighted in the paper, implementing the hub structure noticeably improves test performance. This improvement is particularly significant with smaller training sets, while ESN performance slowly catches up with HubESN as the size of n\_train increases. To confirm that the choice of EC does not affect this trend, we trained HubESN with both smaller (n\_train = 1000-1200) and larger (n\_train = 3000-3200) training sets on the Mackey-Glass time series prediction. The network performance is subsequently assessed on a n\_test=2000 testing set using RMSE. Suppl. Fig. 2 confirms that the choice of EC does not affect this pattern.

\section{Heterogeneity, modularity, and clustering coefficient of hub network}
\subsection{Heterogeneity}
We use the coefficient of variation (CV) of the nodal degree to quantify network heterogeneity [35, 36]. CV measures the variability of the node degrees and is defined as \( CV = \frac{\sigma_{deg}}{\mu_{deg}} \), where \( \sigma_{deg} \) is the standard deviation of the nodal degree and \( \mu_{deg} \) is the mean of the nodal degree. The node degree is defined as the sum of in-degree and out-degree.

\subsection{Modularity}
The modularity is computed by first splitting the network into groups using the Louvain network partition method [37], then measured using the Girvan-Newman modularity measurement algorithm[38]. We use the default setting of \texttt{best\_partition} function in \texttt{community\_louvain} python package to assign community for each node, then use the Girvan-Newman modularity algorithm as defined in Algo. 1 to compute the modularity of the network.

\begin{algorithm}
\caption{Girvan-Newman modularity algorithm}
\begin{algorithmic}
\State $m \gets total\_weight(W)$
\State $k \gets weighted\_degree(W)$
\State $Q \gets 0$
\For {$i \in 0 \to num\_nodes$}
    \For {$j \in 0 \to num\_nodes$}
        \If {$community\_assignments[i] = community\_assignments[j]$}
            \State $Q \gets Q + (W[i, j] - (k[i] \times k[j]) / (2 \times m))$
        \EndIf
    \EndFor
\EndFor
\State \Return $Q / (2 \times m)$
\end{algorithmic}
\end{algorithm}

\subsection{Clustering coefficient}
The clustering coefficient (CC) quantifies the degree to which nodes in a graph cluster together, effectively forming a clique. However, the traditional definition of CC is designed for positive connections and does not accommodate negative edges. Therefore, we removed all negative edges in the graph and used this converted network to estimate the clustering coefficient of the original network. The clustering coefficient of individual nodes is computed using the \texttt{networkx.clustering} function from the \texttt{NetworkX} Python package, and the overall network clustering coefficient is determined by taking the mean CC across all nodes in the network.

\section{The impact of spectral radius and sparsity}
\begin{figure}
  \centering
  \includegraphics[width=0.9\textwidth]{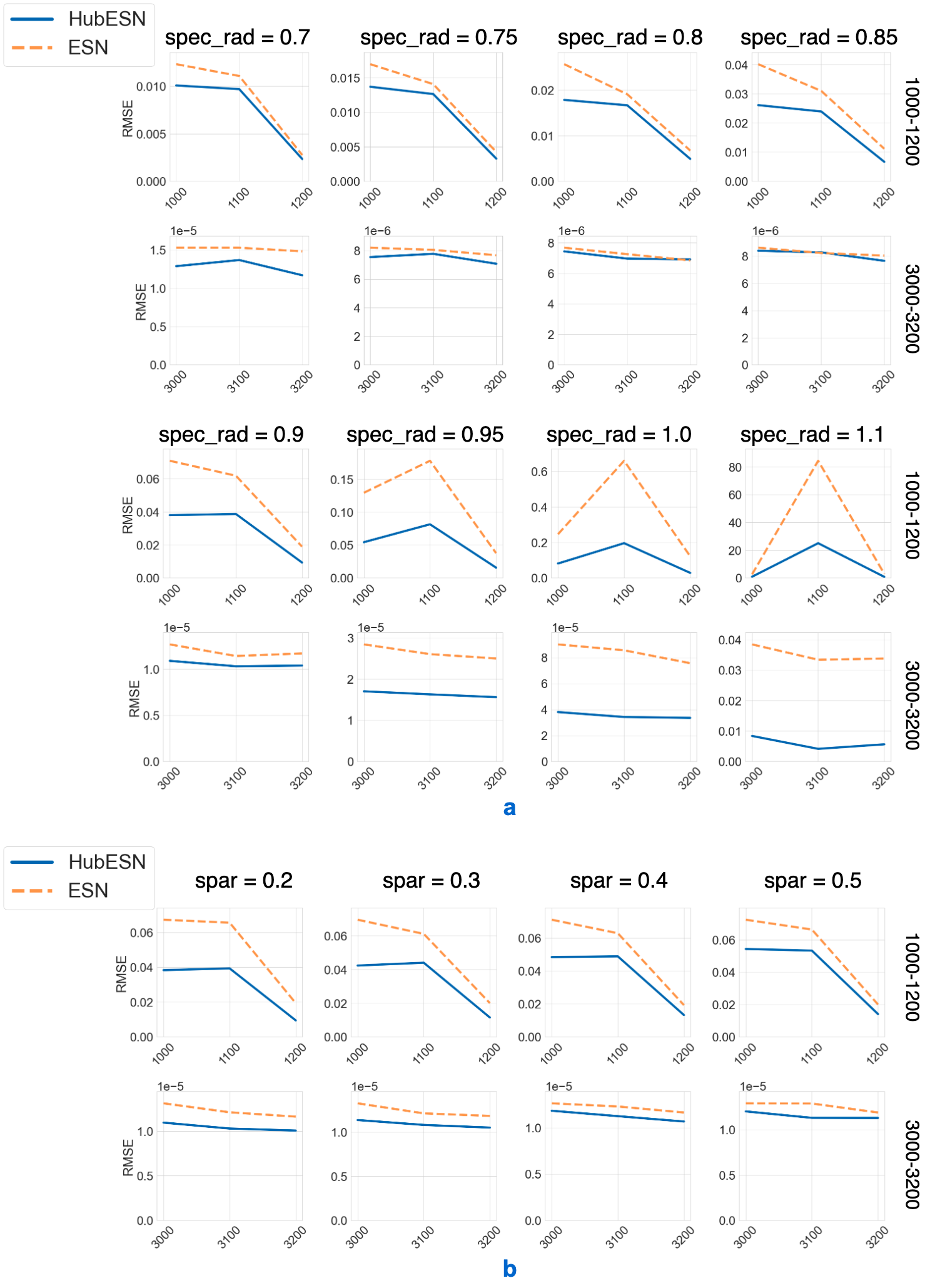}
  \caption{\textbf{a)} Network RMSE on Mackey-Glass prediction under different choices of spectral radius. \textbf{b)} Network RMSE on Mackey-Glass prediction under different sparsity.}
\end{figure}
In our study, we consistently used a spectral radius of 0.9 and a sparsity of 0.2. To ensure these parameter choices did not influence our conclusion, similar to Suppl. section 2, we trained the ESN and HubESN on a smaller and a larger training set and assessed them on a n\_test=2000 testing set. 

As demonstrated in Suppl. Fig. 3a, when the spectral radius is less than 1, it slightly influences the performance of both ESN and HubESN. When n\_train is small, the HubESN consistently outperforms the ESN. As n\_train increases, the performance difference between HubESN and ESN diminishes, which aligns with our findings. On the other hand, when the spec\_rad \( \geq 1 \), the performance of both ESN and HubESN deteriorates significantly, while HubESN has significantly lower RMSE than ESN regardless of the n\_train value.

Furthermore, Suppl. Fig. 3b demonstrates that our choice of sparsity does not significantly alter our conclusions, further aligning with our overall findings.

\section{Implementing ESN, HubESN, and HubESN-rand}
\begin{table}[!h]
  \caption{ESN, HubESN, HubESN-rand parameters}
  \label{model-params}
  \centering
  \begin{tabular}{lll}
    \toprule
    Parameter                     & Value                                          \\
    \midrule
    Input scaling                 & 1                                              \\
    Activation function           & \( tanh( \cdot ) \)                            \\
    \( W^{in} \)                  & \( w_{ij}^{in} ~ \sim \mathcal{U}(-1, 1) \), \( w_{ij}^{in} \in W^{in}\) \\
    \( W^{rec} \)                 & \( w_{ij}^{rec} ~ \sim \mathcal{N}(0, 1/3) \), \( w_{ij}^{rec} \in W^{rec} \) \\
    \( r_{sig} \)                 & 0.1                                            \\
    spec\_rad                     & 0.9                                            \\
    spar                          & 0.2                                            \\
    \bottomrule
  \end{tabular}
\end{table}
\subsection{Model parameters}
In our research, the primary focus is on the influence of the hub structure and hub neurons rather than on parameter optimization. For all the experiments involving ESN, HubESN, and HubESN-rand, we only modify the number of training samples (n\_train) and the location where signals are injected into the hidden layer. Unless specified otherwise, all other parameters remain constant throughout all tests. The specific parameter choices are specified in Table 1.

\subsection{Input and hidden layer initialization}
Both the input layer and the hidden layer are first initialized to a fully connected connectivity matrix with weights following the distribution specified above. The edges in the recurrent weight matrix \( W^{rec} \) are then pruned according to the specified sparsity level. If the network is an ESN, pruning is performed randomly, else if the network is HubESN or HubESN-rand, the pruning is governed by the deletion probability matrix, \( P_{prune} \). Subsequently, the spectral radius is utilized to scale the largest absolute eigenvalue of \( W^{rec} \). Finally, the rows of the input layer will be dropped until \( r_{sig} \) equals 0.1. For ESN and HubESN-rand, the rows in the input layer will be deleted randomly. For HubESN, the rows corresponding to neurons within the lower 90th percentile of nodal degree are eliminated, thereby the signal will only be injected into neurons with top 10\% nodal degrees.

\section{Experimental procedures}
\subsection{Mackey-Glass}
Mackey-Glass time-series is updated as indicated in Eq. 2. In consistent with previous literature, the parameters \( tau \), \( \gamma_{m} \), \( \beta_{m} \), \( k \) are set to \( 17 \), \( 0.1 \), \( 0.2 \), and \( 10 \), respectively. This parameter setting will ensure the system exhibits chaotic behavior. The signal is normalized to \( (-1, 1) \) as we are using \( tanh(\cdot) \) activation function.
\begin{equation}
\frac{dx_{m}}{dt} = \frac{\beta_{m} x(t-\tau)}{1+(x_{m}(t-\tau))^{k}}-\gamma_{m} x_{m}(t)
\end{equation}
Observe that the time series updates using \( x_m(t) \) and the delayed value \( x_m(t-\tau) \). The hidden layer is expected to encode and memorize the delayed signal \( x_m(t-\tau) \). Therefore, at each time step, only \( x_m(t) \) is input to the ESN. After the entire training set \( x_m(t) | t = 0, ..., n \) is input to the network, the readout layer will fit a linear regression on the hidden state \( s(t) | t = 0, ..., n\) to the label \( x_m(t) | t = 1, ..., n+1 \).

\subsection{NARAM10}
NARAM10 is updated as specified in Eq. 3. Similar to the Mackey-Glass task, we expect the hidden layer to encode all previously shown inputs \( x_n(t-i) | i = 1, ... l\). Therefore, at time \( t \) only the most recent input \( u_n(t) \) will be presented to the network, and the network is expected to output \( x_n(t+1) \). However, as the input \( u_n \) for NARMA10 are independent random variables \( u_n(t) \in [0, 0.5] \), the inputs are not normalized as normalizing them will also impact the corresponding label.
\begin{equation}
x_{n}(t) = \alpha_{n} x_{n}(t-1) + \beta_{n} x_{n}(t-1) \sum_{i=1}^{l} x_{n}(t-i) + \gamma_{n} u_{n}(t-l) u_{n}(t-1) + \delta_{n}
\end{equation}

\subsection{MNIST}
Unlike time-series tasks like Mackey-Glass and NARMA10, the MNIST handwritten digit classification is not a time-series prediction task. To make it compatible with the recurrent setting of ESN, we input each 28x28 image into the network column by column over 28 time steps, using one-hot encoding as the label.

In the testing phase, each image produces 28 predictions. The final prediction is derived by majority vote among these predictions. Model accuracy is computed as the proportion of correctly identified labels.

\section{Inverse correlation between readout weights and nodal degree}
\begin{figure}
  \centering
  \includegraphics[width=0.9\textwidth]{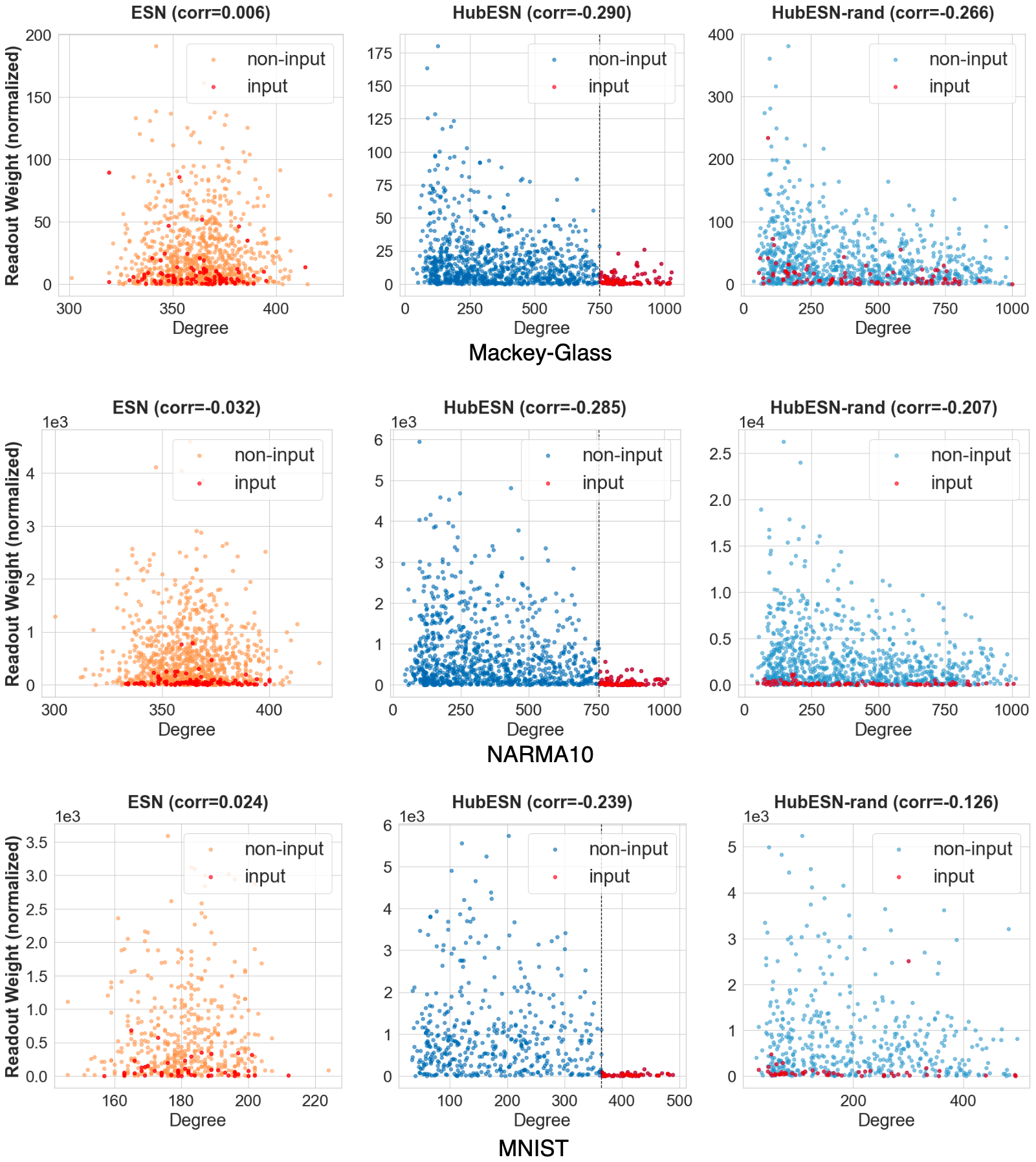}
  \caption{Correlation between node degree and normalized absolute readout weight. In accordance with the experiments in the main paper, we use networks of size 1000 for Mackey-Glass and NARMA10 tasks, and a size of 500 for the MNIST classification task. The training sample size for Mackey-Glass and NARMA10 is 4000, while n\_train=3000 for MNIST. These sizes were chosen based on the point at which network performance ceased to improve.}
\end{figure}
After ESN is trained, the readout layer fits the ESN states across all training time and produces a mapping between ESN states and the output labels. This allows for an additional use of the readout layer, that is its absolute weights can be used to reflect the relative importance of each neuron in prediction. Considering each neuron may be oscillating in different magnitude between \( (-1, 1) \), and lower degree node tends to have lower magnitude. The normalized readout weight of neuron \( n_i \) is computed as follows:
\begin{equation}
w^{norm}_i = |w^{out}_i| \sum_{t=0}^{n\_train} |s_{i,t}|
\end{equation}
Where \( w^{norm}_i \) is the normalized absolute readout weight of \( n_i \), \( w^{out}_i \) is the actual readout weight of \( n_i \) in the readout layer, and \( s_{i,t} \) is the state of \( n_i \) at time \( t \).

Upon obtaining the normalized absolute readout weight for each neuron, we stack the \( w^{norm}_i \) and its corresponding degree \( deg_i \) into two vectors of length \( N \), where \( N \) is the network size. We use \( v_{w} \) and \( v_{deg} \) to represent the two vectors. The correlation between the normalized weight and node degree can be computed using Pearson correlation.

\begin{equation}
r = \frac{\sum(v_{w}-\bar{v_{w}})(v_{deg}-\bar{v_{deg}})}{\sum(v_{w}-\bar{v_{w}})^2 \sum(v_{deg}-\bar{v_{deg}})^2}
\end{equation}

We observed an inverse correlation between the normalized readout weight and nodal degree. This indicates higher nodal degree neurons represent different information than lower nodal degree neurons. The lower nodal degree is preferred during prediction. This pattern is not task-specific, we observed it on all three tasks (Suppl. Fig. 4).
\end{suppl}

\end{document}